# Dense Fully Convolutional Network for Skin Lesion Segmentation

Ebrahim Nasr-Esfahani[a], Shima Rafiei[a], Mohammad H. Jafari[b], Nader Karimi[a],
James S. Wrobel[c], Kayvan Najarian[c,d,e], Shadrokh Samavi[a,f], S.M. Reza Soroushmehr[c,d]

[a] Department of Electrical and Computer Engineering, Isfahan University of Technology, Isfahan 84156-83111, Iran.
[b] Department of Electrical Engineering, University of British Columbia, Vancouver, BC, V6T 1Z4 Canada.
[c] Department of Emergency Medicine, University of Michigan, Ann Arbor, 48109 U.S.A.
[d] Michigan Center for Integrative Research in Critical Care, University of Michigan, Ann Arbor, 48109 U.S.A.
[e] Department of Computational Medicine and Bioinformatics University of Michigan, Ann Arbor, 48109 U.S.A.
[f] Department of Electrical and Computer Engineering, McMaster University, Hamilton, ON, L8S4L8, Canada.

*Abstract*—Lesion segmentation in skin images is an important step in computerized detection of skin cancer. Melanoma is known as one of the most life threatening types of this cancer. Existing methods often fall short of accurately segmenting lesions with fuzzy boarders. In this paper, a new class of fully convolutional network is proposed, with new dense pooling layers for segmentation of lesion regions in non-dermoscopic images. Unlike other existing convolutional networks, this proposed network is designed to produce dense feature maps. This network leads to highly accurate segmentation of lesions. The produced dice score here is 91.6% which outperforms state-of-the-art algorithms in segmentation of skin lesions based on the Dermquest dataset.

*Index Terms*—skin cancer, melanoma, deep neural networks, dense pooling layer.

## 1. Introduction

Computerized diagnosis of skin cancer is of great necessity and interest [1]. About 5.4 million new cases of skin cancer are detected in the USA every year. Most fatal types of skin cancer are melanoma, where 75% of deaths are related to [2]. Recently the incidence pattern of melanoma has shown a rapid increase. The rate of melanoma occurrence has tripled in the past 30 years [3]. In the USA, an estimated 87,110 newly detected cases and an estimated 9,730 melanoma-related deaths occurred only in 2017. A key point in the survival of patients is early detection of malignant skin lesions [4-5]. Patients, with melanoma detected in the early stages, have a 98% of five-year relative chance of survival. The survival rate is shown to be only 18% when melanoma is spread to the other parts of the body, which results in a life expectancy with a median of less than one year [2-3].

The importance and variety of computerized methods for melanoma early detection are reviewed in [6-7]. There exist two main categories of images applied for computerized diagnosis of melanoma: Dermoscopic images, also known as microscopic images, which are captured by a special instrument named dermoscope and non-dermoscopic images, which are captured by conventional digital cameras and smartphones. Dermoscopic images contain more detailed information, while the non-dermoscopic images have the advantage of ease of access. Dermoscopic images are not easily accessible. Authors in [8] indicates that less than 50% of the dermatologists in the United States use dermatoscopy. Jafari et al. [9] have used deep learning to detect melanoma in non-dermoscopic images.

One of the most critical stages in the computerized study of melanoma is the accurate segmentation of skin lesions. Captured images usually contain a lesion, which is surrounded by healthy skin. Proper extraction of the lesion region is critical for assessment of lesion's features. Lesion features include area, border irregularity, shape symmetry, and variation of color. There exist various methods for skin lesions segmentation, based on active contours, region merging [10], and thresholding [11]. Non-dermoscopic images exhibit what is seen by the naked eye. There are some challenges in segmentation of the non-dermoscopic images, such as presence of fuzzy borders between healthy skin and lesion, a variety of textures and colors of lesions, illumination variations in captured images and presence of skin artifacts, like hair. Some researchers have proposed methods for segmentation of non-dermoscopic images [12-16].

Methods of [12], [13], and [14] are based on typical image processing techniques, where vectors of colors and texture features are considered in classifying lesion pixels. In [12] the red channel of RGB images discriminates the lesion from the background. To enhance the segmentation results, the illumination variation by quadratic function is modeled by [13], where the skin pixels are relighted to degrade the shading effects in the RGB image. In [14] a discriminative representation of a 3-channel image is proposed to make a distinction between lesion and background. These methods usually lack accuracy.

Authors of [15] have used sparse texture distribution to discriminate between healthy and malignant skin lesions. This approach yields in noticeable improvement, but it performs poorly in presence of complex lesion and skin patterns.

A relatively successful and widely adopted method is deep learning which has shown an emerging growth in different fields of computer vision [17]. In particular, deep convolutional neural networks have resulted in noticeable advances in different aspects of medical image analysis [18]. Currently deep learning methods are the state-of-the-art framework for various medical imaging challenges, such as classification of mammographic lesions [19], brain lesion segmentation [20], detection of hemorrhage in colored fundus images [21], and bone suppression in X-ray images [22].

The method proposed by [16] is a recent approach for non-dermoscopic image segmentation; the deep convolutional neural networks are applied for lesion segmentation; the pixels into one of the two groups of lesion, and healthy skin are classified by applying both local and global views and it should be added that the method is based on sliding-window, which makes it very time consuming despite the small structure of its network, and in some cases is not fall satisfactory.

The method proposed by [23] is a fully convolutional network for skin lesion segmentation where Jaccard index is applied as a loss function to mitigate the issue of label unbalancing and because this network does not involve low-level features from former layers, for upsampling, it is not appropriate on border regions.

In this paper, a new class of network is proposed named *Dense Fully Convolutional Network* (DFCN). This network generates high-resolution dense feature maps with no need for upsampling similar to sliding-window based methods. These feature maps are generated almost as fast as the fully convolutional networks (FCNs). It can be deduced that this network contain both advantages of dense resolution of sliding-window based methods and the speed of fully convolutional networks.

This network is applied for lesion segmentation and higher dice scores are achieved as compared to another state-of-the-art methods. Briefly, our innovations are:
- Proposal of a new class of fully convolutional networks with high accuracy and speed
- Accurate segmentation of skin lesions for melanoma detection

The rest of the paper is organized such that in Section 2 typical data feeding methods for convolutional neural networks are presented. Then in Section 3 we present our novel proposed architecture for the dense fully convolutional network. Proposed lesion segmentation method is presented in Section 4 and experimental results are shown in Section 5. Concluding remarks are offered in Section 6.

## 2. Typical Data Feeding in Convolutional Neural networks

Typical convolutional neural networks (CNNs) consist of a series of consecutive convolution and pooling layers. The input to the first layer of the CNN is an original image. In each of the convolutional layers, the input is convolved with a predetermined number of kernels to produce feature maps. Moreover, by passing through each pooling layer, the maps size are reduced according to the kernel and the stride size of the pooling layers.

The CNNs architectures, proposed for image segmentation, can be categorized in two: the sliding-window and fully convolutional networks (FCN). In the sliding-window methods each pixel, together with its neighbors, are fed into the network to obtain a label for the pixel. In contrast, an FCN receives the whole image to generate a probability map for the input image. In the following two sub-sections, by using simple structures, we illustrate how the mentioned two groups of methods process the input data. The simple structures consist of two convolution layers and one pooling layer. These structures work on 1-D data strings with 15 elements.

*2.1. Sliding-window method*

In the sliding-window method, a probability is produced for the central pixel of a window and the procedure is repeated for all pixels of the input image. This is accomplished by sliding the window on all pixels of the image. By obtaining probabilities for all pixels and by applying a thresholding process on the obtained probability map, a classification map is yield for the input image.

If a small size window is fed to network, a local view around the pixels will be considered that contains local texture and the precise edge locations. Otherwise, if a large window is defined then a global view is achieved. The global view could extract some general structures and could be beneficial for some applications and could facilitate the classification process. An efficient approach is to combine local and global information by processing both local and global views according to [16] and [24].

In sliding-window, the main issue is the existing redundancy in calculations performed through the network. Since adjacent windows have a considerable number of overlapping pixels, the convolutional and pooling operations would have significant overlaps in assigning labels to adjacent pixels. Another drawback is that the label assignment for each pixel is made separately; therefore, the valuable information about the labels of the surrounding pixels is not involved in the classification procedure. This separate classification in sliding-window methods could result in reduced accuracy in some applications.

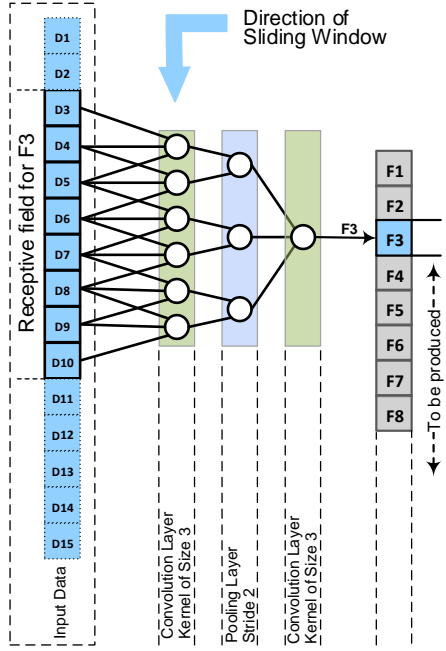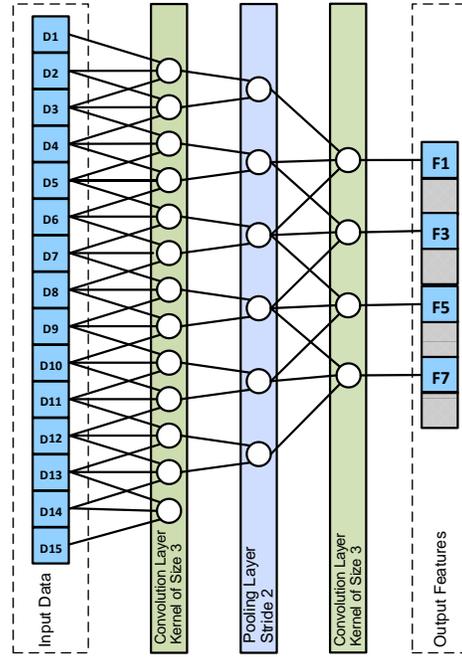

Fig. 1. Generation of output features in a sliding-window structure.

Fig. 2. Generation of output features in a fully convolutional network structure.

As shown in Fig. (1), a sliding-window of size 8, slids on the input data, where 8 features of F1-F8 are obtained. In other words, a single feature is obtained from its corresponding receptive field, or input window of size 8.

*2.2. Fully convolutional method*

FCN is constructed by replacing the fully connected layers with a series of deconvolution layers where the size of the input image could vary and it no longer depends on the network's structure.

The general structure of FCN consists of two phases of convolutional encoding and decoding. In FCN, contrary to the sliding-window approach, the image as a whole is fed to the network. At the encoder phase, low-resolution feature maps are obtained by passing the entire image through convolutional and pooling layers. Then the convolutional decoder deconvolves the produced low-resolution feature map into a map of almost the same size as the input image. Hence, the network generates the segmentation probability map of the whole image. Deconvolution is usually done by one step interpolation [25] or by a series of deconvolution and upsampling layers [26-27].

The accuracy of the FCN highly depends on the accurate deconvolution of the low-resolution feature maps. In [27] and [28], the indices of pooled features are stored in pooling layers of the convolutional encoder. The stored locations and their indices are then applied in convolutional decoder phase to improve the accuracy for generating the final map. A U-shape network structure for medical image segmentation is proposed in [26] where in each one of the deconvolution layers, the corresponding feature maps from the encoder stage are concatenated with the upsampled maps in order to improve deconvolution accuracy.

By applying the FCN method, as can be seen in Fig. (2), the entire input data is convolved in one step. Then after pooling and second convolutional layer, the 4 features, namely F1, F3, F5, and F7 are obtained. Each one of the output feature corresponds to a specific input window, while, some output features, such as F2, F4, F6, and F8 are not generated due to the presence of the max pooling layer with stride 2.

Some of the overlapped input windows would be ignored by the pooling layer, with stride 2. As a result, the corresponding output features would be missed. These missing features could be reconstructed by deconvolution layers with the possibility of imprecise output.

Compared to sliding-window method, in FCN by reducing the redundancy in network's operations for adjacent pixels, the network's training and testing speed is significantly increased. Meanwhile, the added deconvolution decoder layers increase the number of required parameters and required calculations, which could impair the network's learning process.

### 3. Dense Fully Convolutional Network

By introducing an innovative pooling method, Dense Fully Convolutional Network (DFCN), the computational redundancies in calculations of adjacent windows in the sliding-window approaches are avoided. The DFCN's encoder directly outputs high-resolution feature maps with almost the same size as the input image; therefore, in DFCN the need for the deconvolution phase is eliminated. Absence of the deconvolution phase allows DFCN to have a high speed like FCNs. On the other hand, DFCN applies an innovative pooling method that eliminates the deconvolution steps in FCNs.

In FCNs, a pooling layer divides a 2-D input into a set of non-overlapping rectangular sub-regions, which are commonly 2×2. The pooling layer performs a non-linear function, which is commonly a max pooling operation with a stride of 2. This means the pooling layer reduces the spatial size of feature maps for the subsequent layers. Hence, max pooling strongly reduces the number of parameters for the rest of the network. Moreover each neuron in the subsequent layer, which follows the pooling layer, would have larger field of view. This larger field of view, which receives information from a larger neighborhood of pixels, is subject to the pooling effect, making the detection process more powerful. Nevertheless, this approach causes missing of some features and production of low-resolution maps, which need reconstruction by deconvolution layers.

In DFCN, instead of using non-overlapping rectangular sub-regions, the pooling layers consider all possible overlapped sub-regions. Unlike FCN, in this process through DFCN no feature is missed.

To illustrate how this DFCN works, suppose we have a simple 1-D network as shown in Fig. (3). Similar to the networks of Figs. (1 and 2), this network consists of two convolution layers, both with kernels of size 3, and a pooling layer, with a kernel of size 2 and stride of 2. Unlike the networks of Figs. (1 and 2), the pooling layer in Fig. (3) is a dense-pooling layer. The shown network is in 1-D, while it can easily be generalized to higher dimensional spaces.

As shown in Fig. (3), the input data, after convolution in the first layer, would be fed into the dense-pooling layer. In the dense pooling layer, there would be a pooling with a stride of 1. The even and odd features are separated into two upper and lower categories. The top category consist of features that would be produced by the FCN of Fig. (2), and the lower category consists of features that are ignored by the FCN.

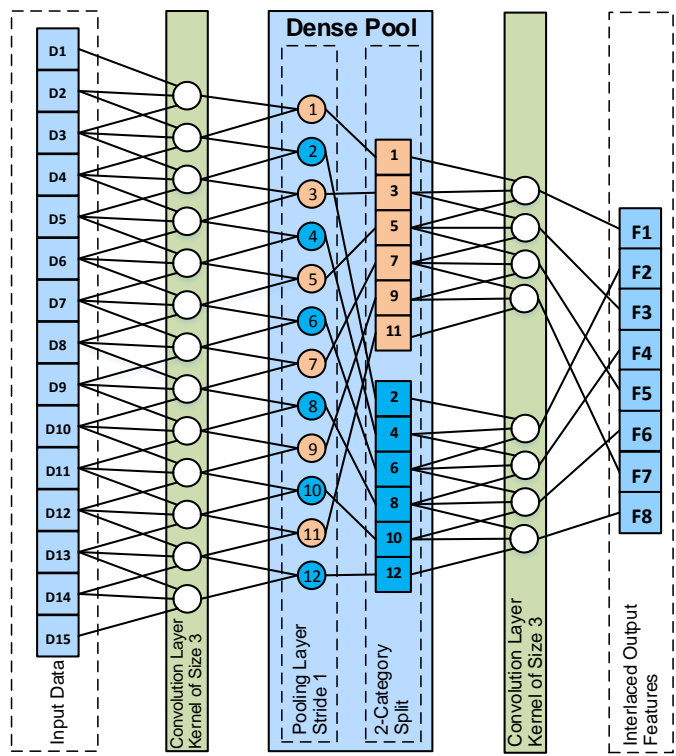

Fig. 3. General structure of a 1-D dense fully convolutional network (DFCN).

Afterward, both categories are fed into the next convolutional layer and would be convolved with the same kernel sharing weights. At the final stage of the network, all features that are in different categories are interlaced and they placed in their correct locations. As can be seen, the DFCN's output would be exactly the same as the sliding-window method due to the fact that the dense pooling layer does not ignore any receptive field. Consequently, DFCN produces a high-resolution feature map without the need for any deconvolution step. Hence, a large number of parameters are eliminated, which results in a network that is easier to train.

Thus, DFCN produces the high-resolution feature maps of sliding-window based networks and high speed of FCNs. The speedup of DFCN is achieved because all of the overlapping computations, which exist in the sliding-window methods, are avoided by dense pooling.

In all of the above mentioned networks, the structures and the number of parameters are the same, while the number of operations required to produce these features are different. In Table 1 the number of multiplications and the number of addition operations are shown for each network. It should be noted that in a convolutional layer, with a kernel size of 3, each neuron has to perform 3 multiplications, the outcome of which should be added together. Hence, the total number of multiplications is equal to 3 times the number of neurons. Also, the total number of additions of a layer is equal to the number of neurons in that layer. The number of features achieved by sliding-window is identical to that of DFCN. Nevertheless, since we removed redundancies in calculations, the number of operations used by the sliding-window is almost 3 times more than that of the DFCN. Moreover, the number of operations in DFCN is almost 20 percent more than the number of operations in FCN. However the features produced by DFCN are more precise and there are no missing feature. Hence, unlike FCN, there is no need for additional deconvolution layers for reconstruction of the missing features.

Table 1. Computational complexity based on the number of addition and multiplication operations

| Method | # of Multiplications | # of Additions |
|---|---|---|
| Sliding-Window | 168 | 70 |
| FCN | 51 | 17 |
| DFCN | 63 | 21 |

To generalize, when a 2-D image is fed into the DFCN, the first dense pooling layer, with stride $S$, at the first step, pools each of its input feature maps with stride one. Then it splits set of one-stride maps. This means that each single category is split into $S^2$ categories. All these generated categories are of equal size with no overlap. Also, the union of these categories is the same as the set of one-stride maps. In the same manner, each subsequent dense pooling layer would split each of its input categories into $S^2$ categories. As a result, the number of categories is increased over the network.

The splitting procedure is explained by Eq. (1). In Eq. (1), for each category, each feature map, $F$, is split and placed into $S^2$ categories. In Eq. (1), $m$ and $n$ are the row and column indices of the new categories, while $x$ and $y$ indicate the spatial location in each category.

$$F^{mn}(\lfloor x/s \rfloor, \lfloor y/s \rfloor) = F(\lfloor x/s \rfloor + m, \lfloor y/s \rfloor + n);$$
$$\forall (m,n) \in \{0,1 \dots, S-1\} \tag{1}$$

### 4. Proposed Lesion segmentation method

We have developed a network based on DFCN structure for segmentation of skin lesions in non-dermoscopic images. In this section, the structure of the network is described, followed by the strategies applied for improving the training process and increasing the effectiveness of the network.

*4.1. Network architecture*

The DFCN structure of Fig. (4) converts an RGB image into a posterior probability map, which will eventually produce the segmentation of the image. The input that is fed into DFCN may be the whole image or a part of the image. The output of DFCN is a probability map, indicating the probability of each pixel's membership in the lesion or the background region. In this network, the feature maps are produced through 5 paths of different lengths. The shorter paths generate low level features, like border information. The longer paths generate high level features, like structural shapes. The final results of these five paths are interlaced to place each feature at its correct location. Then through two final convolutional layers, the network concatenates these interlaced feature maps to generate a final probability map.

In this proposed DFCN, all dense pooling layers are max pooling with 2×2 kernels and stride 2. Hence, each dense pooling would split each category that exists in the previous layer into 4 categories. For this purpose, at the first step, each dense pooling layer would pool every category with stride one according to Eq. (2). In this equitation, for each category, variable $F_j^l$ is the $j^{th}$ feature map of the $l^{th}$ layer. At the second step, since we are using stride 2, the dense pooling would split each pooled category into 4 categories.

$$F_j^l(x,y) = max\left\{\{F_j^{l-1}(x+m, y+n); \forall (m,n) \in \{0,1\}\right\} \tag{2}$$

At the convolutional layer each category would be convolved separately according to Eq. (3), where: $N^l$ is the number of feature maps of the category in the $l^{th}$ layer. Also, $w_{ij}^l$ is the $i^{th}$ kernel applied to the $j^{th}$ feature map of the category and $b_j^l$ is the $j^{th}$ bias for this category. It should be noted that there are no connections between categories, while, in all convolutional layers the kernel weights are shared among all categories. These categories would not join each other and are kept separated until reaching the interlace layer. At the interlace layer, all categories would be interleaved to achieve the output with almost the same size as the whole image.

$$F_j^l = \sum_{i=1}^{N^l}(F_i^{l-1} * w_{ij}^l) + b_j^l 1_{N^l} \qquad (3)$$

For achieving further precision in the segmentation results, we crop and convolve feature maps from various layers to combine both high and low-level information. Subsequent interlace layers would relocate each feature to its correct location. This leads to set of high-resolution feature maps. Then the network concatenates these dense feature maps, which will be followed by two convolution layers. At the final stage of the network, the corresponding probability map would be produced, using softmax function of Eq. (4).

$$p_k(x,y) = \exp(a_k(x,y)) / \sum_{k'=1}^{C} \exp(a_{k'}(x,y)) \qquad (4)$$

where $a_k(x,y)$ and $p_k(x,y)$ are respectively the $k^{th}$ activation output and its probability of belonging to the $k^{th}$ class at the location $(x,y)$ and $C$ is the number of classes. For lesion segmentation $C$ is 2, indicating the existence of both classes of lesion and background.

*4.2. DFCN training*

The number of non-dermoscopic images in the training set as compared to the number of network parameters, is small. Thus to enhance DFCN performance and overcome a possible over-fitting problem caused lack of the training set, the following five strategies are implemented:

*4.2.1. Sub-Imaging*

Similar to FCN, this DFCN does not contain any fully connected layer, consequently, in the test or training phases, the entire or a part of an image with different size can be fed into the network. In this new approach, the network is trained with sub-images of 155×155 size, which are cropped from the original image. Since most portion of a non-dermoscopic image belong to the background area, pure random selection of the sub-images would generate an unbalanced training set. This unbalance training set leads to a reduction of the network performance. Therefore, we select sub-images such that 25% of the sub-image centers are located in the lesion area and 25% are in the background area. The remaining 50% of the sub-image centers are within the lesion border areas, which are computed using the morphological procedure of Eq. (5).

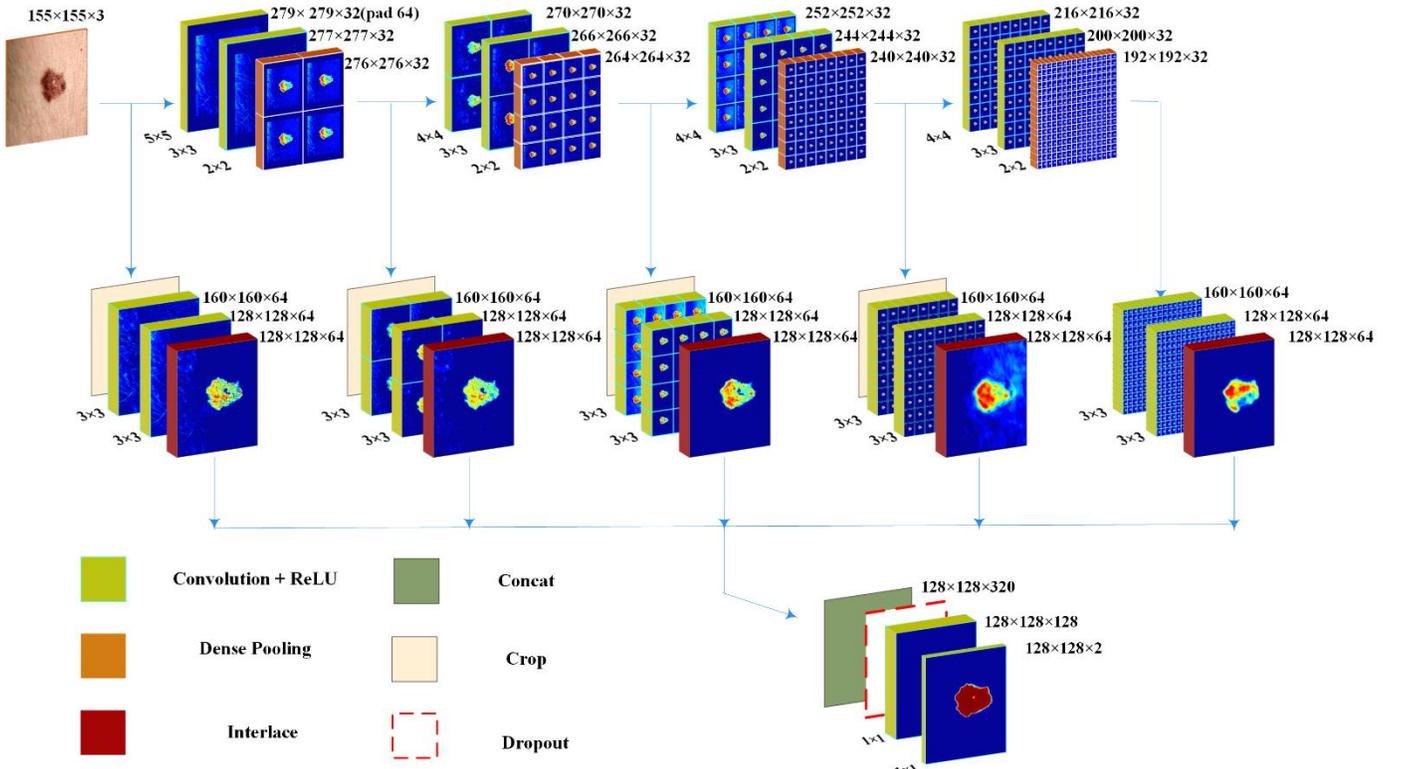

Fig. 4. The structure of dense fully convolutional network (DFCN) for lesion segmentation. The upper annotation each layer indicates the output size of each layer while the lower annotation demonstrates the kernel size.

$$border = (GT \oplus SE) - (GT \ominus SE) \tag{5}$$

where $\oplus$ and $\ominus$ are the dilation and erosion operators, respectively. Also, the $GT$ is the segmentation ground truth and $SE$ is a disk structuring-element of size 20.

*4.2.2. Pixel weighting*

The last layer of this network is the soft-max layer, where the probability of each pixel's membership to the lesion or background areas is determined. Our DFCN has been trained by the cross-entropy loss function that is shown in Eq. (6).

$$E = - \sum_{(x,y)\epsilon\mathbb{Z}^2} w(x,y)\log(p_c(x,y)) \tag{6}$$

where $p_c(x,y)$ is the probability of the pixel in location $(x,y)$ belonging to the lesion area, which is obtained from the softmax function. Also, $w(x,y)$ is a 2-D weighting map that is calculated by Eq. (7). Pixels that are closer to the boundary would gain higher weights through Eq. (7). In other words, $w(x,y)$ in the training phase would increase the importance of pixels that are closer to the lesion boundaries. Labeling such pixels is relatively hard since borders are usually vague.

In Eq. (7), $d(x,y)$ is the distance between the pixel and the lesion boundary, where, $w_0$ and $\sigma$ are selected to be 20 and 30, respectively.

$$w(x,y) = 1 + w_0 \cdot \exp(-\frac{d(x,y)}{2\sigma^2}) \tag{7}$$

*4.2.3. Adam stochastic optimization*

The momentum optimization [30] is introduced to assist accelerate the stochastic gradient descent (SGD) [29]. This is done by reducing oscillations and leading SGD to a relevant direction, at the expense of defining an additional hyperparameter. For this purpose, the Adam algorithm known as adaptive moments [31] is applied. It is found that Adam optimization is relatively robust for the choice of hyper parameters in this implementation. We have set the learning rate to 0.0001 and adopted Adam's default parameters for the first and second moments.

*4.2.4. Image augmentation*

To compensate for the lack of training set which contains only 126 images, we need to augment the dataset. Choosing a large number of sub-images from a single image will generate overlapped information. Hence we artificially augment the training set. This would mitigate the over-fitting issue and could lead to well-trained weights. For this purpose, we have rotated each image from 0 to 360 degrees in 10 steps. After each rotation we select 70 sub-images. Each chosen sub-image is flipped horizontally with a probability of 0.5, and then it is flipped again vertically with the same probability.

*4.2.5. Dropout*

In our proposed network, the large number of parameters might result in over-fitting issue and the failure in the network's training procedure. To overcome this issue, by using dropout regularization technique [32], in a given layer with probability $\rho$, a subset of neurons will be dropped out as inactive neurons. These inactivated neurons have no contribution in the feedforward and back propagation processes. In such a way, since active neurons cannot rely on the dropped-out neurons they are forced to learn more robust features independently. Therefore, the network will be well-trained even when it has limited data. In this paper, as shown in Fig. (4), we use dropout with $\rho = 0.5$ after the concat layer.

## 5. Result and Discussion

In this section, we have described the implementation details such as the used dataset and the network settings. We also present both quantitative and qualitative performance of the proposed method. Our implementation of the proposed DFCN network is available online[1].

*5.1. Implementation details*

Our DFCN was tested on images from Dermquest, an online publically available dataset, accompanied by the segmentation ground truth [33]. This dataset includes 137 non-dermoscopic images. Nevertheless we omitted 11 images that are recently added to the dataset. This omission is done for fair comparison with methods which have reported their results on the previous dataset. The 126 selected images consist of 66 melanoma and 60 non-melanoma cases. The images were randomly split into four distinct folds with almost equal number of images. Images were tested based on leave-one-out cross-validation technique, through which one fold is selected for testing, two folds for training and one fold for validation at a time. In other words, approximately 63 images of this data-set are used for training, 31 images for validation and the rest for testing.

---
[1] https://www.github.com/ebrahimnasr/DFCN/

A set of 700 sub-images with the size of 155×155 is generated from each image. A random vertical, horizontal fillping and rotating has produced these sub-images. Therefore, our DFCN training was performed on 44100 (i.e. $126 \times 700 \times 1/2$) images. The network weights were initialized by Xavier method [34] while the biases in all layers were set to 0. The early stopping technique is also applied to avoid over-fitting in every iteration. The proposed method is implemented in python and Caffe [35], on a computer with an Intel Core i7-4790K processor, 32 GB of RAM, and an NVIDIA GeForce GTX Titan X GPU card.

*5.2. Quantitative Evaluation*

The DFCN outputs a probability map, in which the probability of belonging to lesion or background is determined for each pixel. By setting the threshold to 0.5, this map is converted to a binary mask. Our segmentation results have been compared with those of [10], [12-16], [23], [25], and [26]. These algorithms are tested on the Dermquest database. L-SRM [10], Otsu-R [12], Otsu-RGB [13], Otsu-PCA [14] and TDLS [15] are non-deep learning methods. The method proposed by Jafari [16] is a sliding-window based deep network. The method of [23] is fully convolutional network based on Jaccard loss function. The algorithm of FCN-8s [25] has a better performance among the FCN family methods. U-net [26] is used for segmentation of medical images. Both FCN-8s and U-net are based on fully convolutional structures. In an experiment, we examined the network without the proposed interlace layer. Our DFCN with no interlace layer achieved the result reported in Table 2. This network achieved less accurate results. This is due to the fact that the penultimate convolutional layer, which has to convolve concatenated features, receives its input features from non-corresponding locations. Previously, the interlace layer was responsible for placing features in their correct positions. Comparative results are presented in Table 2. As shown in this table, our network yields 91.6% dice score, a criterion with emphasis only on the correct segmentation of the lesion area. Consequently we outperform all mentioned methods on accuracy, dice score, Jaccard index and border error criteria.

Table 2. Quantitative comparison of lesion segmentation methods applied to the Dermquest database images. Best results are bolded.

| Segmentation algorithms | Method type | Dice Score (%) | Accuracy (%) | Sensitivity (%) | Specificity (%) | Border error | Jaccard index (%) |
|---|---|---|---|---|---|---|---|
| L-SRM [10] | Non-CNN | 82.2 | 92.3 | 89.4 | 92.7 | 57.7 | 70.0 |
| Otsu-R [12] | Non-CNN | 73.7 | 84.9 | 87.3 | 85.4 | 55.4 | 64.9 |
| Otsu-RGB [13] | Non-CNN | 69.3 | 80.2 | 93.6 | 80.3 | 70.0 | 59.1 |
| Otsu-PCA [14] | Non-CNN | 83.3 | 98.1 | 79.6 | 99.6 | 40.8 | 75.2 |
| TDLS [15] | Non-CNN | 82.8 | 98.3 | 91.2 | 99.0 | 39.7 | 71.5 |
| Jafari [16] | CNN-Sliding Window | 83.1 | 98.7 | **95.2** | 99.0 | 23.0 | 81.2 |
| FCN-8s [25] | CNN-based | 89.7 | **98.9** | 90.0 | 99.5 | 19.0 | 82.9 |
| U-net [26] | CNN-based | 88.7 | 98.7 | 91.5 | 99.5 | 22.4 | 81.4 |
| Yuan [23] | CNN-based | 89.3 | 98.7 | 91.6 | 99.6 | 20.7 | 82.0 |
| DFCN-No Interlace | CNN-based | 81.62 | 98.0 | 78.0 | **99.8** | 28.4 | 73.0 |
| DFCN | CNN-based | **91.6** | **98.9** | 92.4 | 99.6 | **16.5** | **85.2** |

*5.3. Qualitative Evaluation*

In Fig. (5) visual quality of our proposed method is compared with 4 other methods of Jafari [16], FCN-8s [25], U-net [26] and TDLS [15]. At the first row of Fig. (5) four challenging images and their ground truths are shown. In comparison to other methods, our method segments lesion region more precisely. This is due to DFCN generating accurate probability maps. Other methods suffer from probability maps with artifacts. For example, in Fig. 5(b-1), (c-2), and (d-4) these artifacts are present.

In addition, the segmentation result of FCN-8s, U-net and TDLS in Fig. 5(b-2) and (e-3) are not accurate on boundaries of lesions and they misclassified the pixels that are near the borders. In fact our method generated proper probability map on lesion borders that leads to precise boundaries. In Fig. 5(a-3) the dermatologist considered the pale reddish inflammation together with the dark part as the lesion. Although, similar to other methods, we did not completely detect the pale reddish inflammation area, only our method and FCN-8s partially detected the mentioned inflammation region.

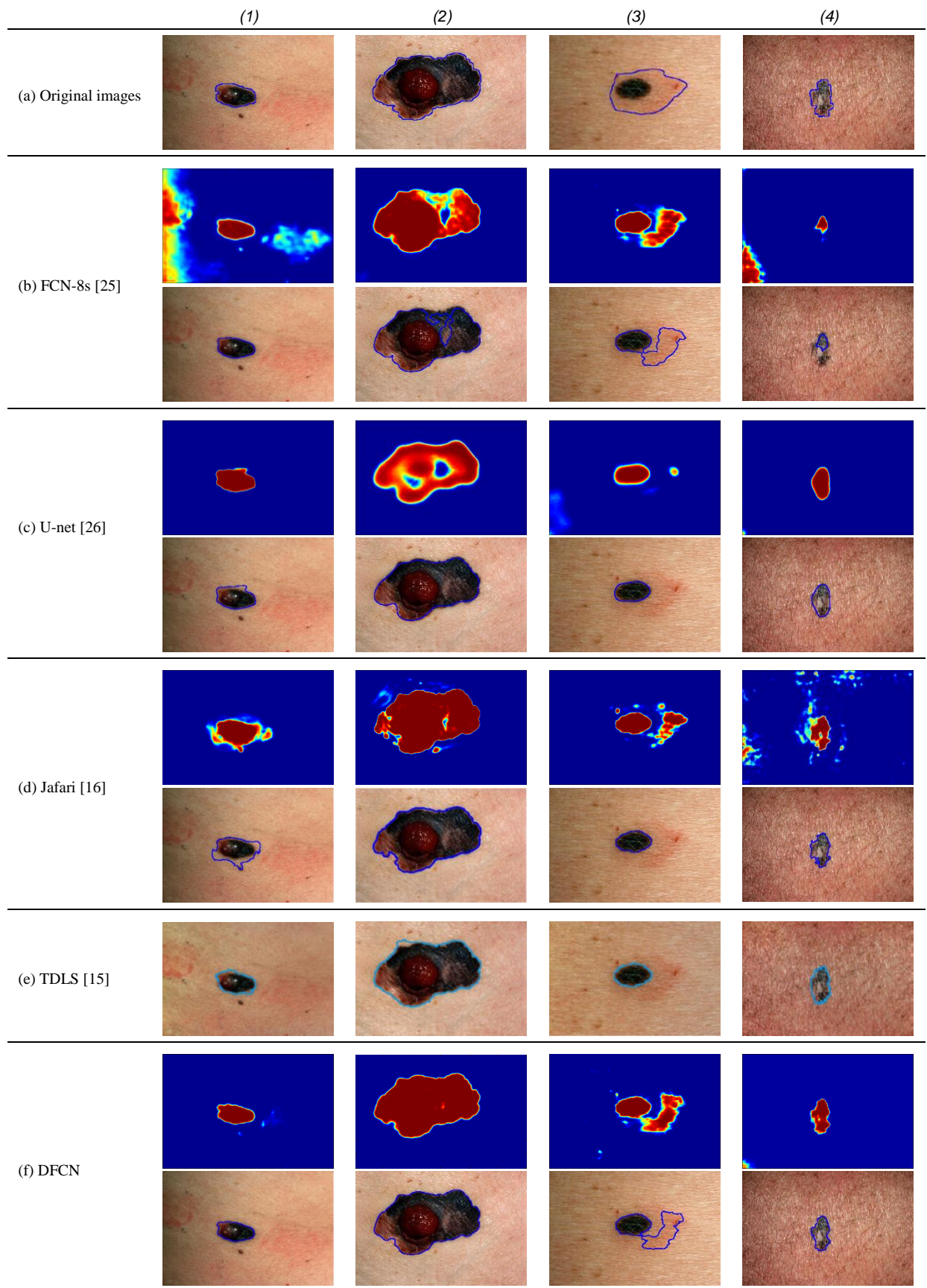

Fig.5. (a) Four challenging images with ground truths shown by blue borders, (b) probability maps and segmentation results of FCN-8s [25], (c) probability maps and segmentation results of U-net [26], (d) probability maps and segmentation results of Jafari [16], (e) segmentation results of the non-deep learning method of TDLS [15], (f) probability maps and segmentation results of the proposed DFCN.

## 6. Conclusion

In this paper, we proposed a new class of fully convolutional networks called DFCN. We proposed dense pooling layers which preserve their features rather than losing them by greater-than-one strides. This preservation of features leads to dense resolution feature maps. Therefore our DFCN eliminates the need for a decoder phase to reconstruct the missing features. Meanwhile, our network is as fast as FCNs methods. The proposed DFCN produced a dice score of 91.6% on Dermquest database. We experimentally showed that the proposed network could outperform state-of-the-art methods in skin lesion segmentation of non-dermoscopic images. Our comparisons included well-known deep-based methods such as FCNs and U-net.